# The MCC-F1 curve: a performance evaluation technique for binary classification

Chang Cao[1]

University of Toronto

& Princess Margaret Cancer Centre

Davide Chicco[1,2]

Princess Margaret Cancer Centre

Michael M. Hoffman[*]

Princess Margaret Cancer Centre

& University of Toronto

& Vector Institute

17th June, 2020

**Abstract**

Many fields use the receiver operating characteristic (ROC) curve and the precision-recall (PR) curve as standard evaluations of binary classification methods. Analysis of ROC and PR, however, often gives misleading and inflated performance evaluations, especially with an imbalanced ground truth. Here, we demonstrate the problems with ROC and PR analysis through simulations, and propose the MCC-$F_1$ curve to address these drawbacks. The MCC-$F_1$ curve combines two informative single-threshold metrics, Matthews correlation coefficient (MCC) and the $F_1$ score. The MCC-$F_1$ curve more clearly differentiates good and bad classifiers, even with imbalanced ground truths. We also introduce the MCC-$F_1$ metric, which provides a single value that integrates many aspects of classifier performance across the whole range of classification thresholds. Finally, we provide an R package that plots MCC-$F_1$ curves and calculates related metrics.

*Keywords*: binary classifier; confusion matrix; biostatistics; machine learning; data mining; contingency table; binary classification evaluation; Matthews correlation coefficient; $F_1$ score; ROC curve; precision-recall curve.

## 1 Introduction

### 1.1 Classifier evaluation methods

The receiver operating characteristic (ROC) curve [1] and the precision-recall (PR) curve [2] provide commonly-used methods to assess and compare binary classifier performance. A *binary classifier* transforms each element $x_i$ in an input dataset into a positive or negative prediction $\hat{y}_i$. *Scoring classifiers* do this by producing a real-valued *prediction score* $f(x_i)$ for each element, and then assigning positive predictions ($\hat{y}_i = 1$) when the score exceeds some threshold $\tau$, or negative predictions ($\hat{y}_i = 0$) otherwise [3]. By comparing a list of predictions with corresponding binary *ground truth* labels $y_i$, one may generate a *confusion matrix*. A confusion matrix is a 2×2 contingency table that summarizes classifier performance at a particular threshold $\tau$. When there exists no predefined threshold $\tau$, it is common practice to compute confusion matrices for a range of thresholds. Often, researchers will generate a confusion matrix for each prediction score $f(x_i)$ produced from the input dataset. One may then compute other statistics to summarize the confusion matrix, emphasizing different aspects. With two different summary statistics, one can plot their values on a two-dimensional curve over varying values of $\tau$. The curves most commonly used for this purpose are the ROC curve and the PR curve.

[*]corresponding author: Michael M. Hoffman, michael.hoffman@utoronto.ca
[1] co-first author
[2] currently at the Krembil Research Institute

|  |  | predicted positive $\hat{y}_i = 1$ | predicted negative $\hat{y}_i = 0$ |
|---|---|---|---|
| actual positive | $y_i = 1$ | true positives (TP) | false negatives (FN) |
| actual negative | $y_i = 0$ | false positives (FP) | true negatives (TN) |

**Table 1: Confusion matrix cells**. For a binary classifier, the four cells of the confusion matrix quantify the frequency of every combination of predicted class $\hat{y}_i$ and actual class $y_i$.

For certain problems, researchers face the challenge of evaluating classification with *imbalanced ground truth*, where the number of negative labels and positive labels differ substantially. Imbalanced ground truth occurs often in medicine and public health, where the number of healthy controls might greatly exceed the number of disease cases. Genomics also frequently poses imbalanced ground truth problems, where the proportion of the genome deemed "interesting" represents a small fraction of the total. Many have identified how ROC analysis on imbalanced data can mislead [1, 4, 5, 6, 7]. While some hold out PR analysis as a solution to this problem [4, 5], we show below that it can prove ineffective on some imbalanced datasets.

Researchers have proposed other curves to overcome the problems of existing assessment methods. The Cost Curve [8] proves useful in testing various functions of misclassification cost, but gives less importance to correct predictions. The concentrated ROC (CROC) [9] addresses *early retrieval* problems. Early retrieval problems emphasize the correctness of a small number of highest-ranked predictions, often where one can only put a limited number of predictions to practical use. For example, only the very top of a ranked list of predictions proves of interest in web search or drug lead identification. Like ROC analysis, however, CROC analysis cannot differentiate the performance of a classifier under different class imbalances [4]. The total operating characteristic (TOC) [10] provides more information than the ROC curve. The complexity of its interpretation may have prevented its wider use in the scientific community. The Precision-Recall-Gain curve [11] enhances precision and recall statistics through assessment of *positive gain*—how much the classifier's performance surpasses an always-positive classifier. While often an improvement over PR analysis, Precision-Recall-Gain analysis still focuses only on correct predictions of positive cases and excludes direct evaluation on negative cases. The *partial* ROC curves [12, 13] present some advantages with respect to the traditional ROC curves, but fail to consider the complete dataset in their analyses.

Because of these flaws and limitations found in these performance analyses, we decided to design a new curve, able to provide a clear, balanced evaluation of any confusion matrix. Here, we present a novel statistical method to assess the performance of classifiers over multiple thresholds, based on the Matthews correlation coefficient (MCC) [14] and the $F_1$ score [15]: the MCC-$F_1$ curve. Our method shows visually how classifier choice and parameters affect all four cells of the confusion matrix, while considering the ground truth class prevalence.

## 1.2 Summarizing the confusion matrix

The confusion matrix's four cells (Table 1) specify separately its performance on positive-labeled elements (actual positives) and negative-labeled elements (actual negatives). Actual positives correctly classified as positive are *true positives* (TP) and those wrongly classified as false are *false positives* (FP). Actual negatives correctly classified as negative are *true negatives* (TN) and those wrongly classified as false are *false negatives* (FN).

For example, suppose we have 100 positive items and 100 negative items. We successfully predict 60 positives to be positive, and 50 negatives to be negative. We mistakenly predict 40 positives to be negative, and 50 negatives to be positive. A confusion matrix that summarizes the frequency of all four outcomes, would have TP = 60, FN = 40, FP = 50, and TN = 50.

| type | metric | TP | FP | TN | FN | multi-threshold | worst value | best value | equation |
|---|---|---|---|---|---|---|---|---|---|
| single-aspect | recall (TPR) | + | | | − | | 0 | 1 | Eqn 1 |
| | false positive rate (FPR) | | + | − | | | 1 | 0 | Eqn 2 |
| | precision | + | − | | | | 0 | 1 | Eqn 3 |
| total-summary | accuracy | + | − | + | − | | 0 | 1 | Eqn 4 |
| | balanced accuracy | + | − | + | − | | 0 | 1 | Eqn 5 |
| | MCC | + | − | + | − | | −1 | +1 | Eqn 6 |
| | unit-normalized MCC | + | − | + | − | | 0 | 1 | Eqn 8 |
| | $F_1$ score | + | − | | − | | 0 | 1 | Eqn 7 |
| multi-threshold integration | AUROC | + | − | + | − | ✓ | 0 | 1 | |
| | AUPR | + | − | | − | ✓ | 0 | 1 | |
| | MCC-$F_1$ metric | + | − | + | − | ✓ | 0 | 1 | Eqn 15 |

**Table 2: Taxonomy of binary classifier performance metrics.** Single-aspect metrics capture only one row or column of the confusion matrix. Total-summary metrics integrate multiple aspects of the confusion matrix. Multi-threshold integration metrics summarize a classifier's performance across multiple thresholds. "+" (or "−") means the metric increases (or decreases) with an increase in the number of true positives (TP), false positives (FP), true negatives (TN), or false negatives (FN). Blanks under TP, FP, TN, or FN indicate that the corresponding metric does not consider the corresponding cell of the confusion matrix.

While a confusion matrix shows the raw counts of all four outcomes, it is difficult to interpret classifier performance with these counts alone. Instead, we can employ metrics that produce a single value by combining multiple quantities from the confusion matrix (Table 2). *Single-aspect metrics* capture only one row or column of the confusion matrix, and include

$$\text{recall} = \text{TPR} = \frac{\text{TP}}{\text{TP} + \text{FN}} \quad (1)$$

(worst value = 0; best value = 1),

$$\text{false positive rate (FPR)} = \frac{\text{FP}}{\text{FP} + \text{TN}} \quad (2)$$

(worst value = 1; best value = 0), and

$$\text{precision} = \frac{\text{TP}}{\text{TP} + \text{FP}} \quad (3)$$

(worst value = 0; best value = 1).

The PR curve plots precision (Eqn 3) against recall (true positive rate (TPR)) (Eqn 1), and the ROC curve plots recall (TPR) against FPR (Eqn 2). Each of the plotted variables is a single-aspect matrix emphasizing different aspects of the confusion matrix. Precision measures exactness—how many predicted positives were actual positives. Recall, instead, measures completeness—how many actual positives were predicted positives [16]. FPR measures how many examples of the negative class are mislabeled. Precision, recall, and FPR capture the confusion matrix only partially. For example, neither true negatives nor false positives affect recall, and a classifier that simply predicts that all elements are positive has high recall [17].

One can use *total-summary metrics* to capture more aspects of the confusion matrix in a single quantity with various trade-offs. The simplest way to combine all cells of the confusion matrix is *accuracy*, which divides the number of correct predictions by the total number of predictions:

$$\text{accuracy} = \frac{\text{TP} + \text{TN}}{\text{TP} + \text{FN} + \text{TN} + \text{FP}} \quad (4)$$

(worst value = 0; best value = 1).

Accuracy, however, leads to an optimistic estimate when used to test a biased classifier on an imbalanced dataset [18]. For example, if we took a dataset with 90% actual negatives and 10% actual positives and predicted all elements as negative, we would have an over-optimistic inflated accuracy = 0.9.

*Balanced accuracy* [18] addresses some of the above issues above by comparing true positives with actual positives separately from comparing true negatives with actual negatives:

$$\text{balanced accuracy} = \frac{1}{2} \cdot \left( \frac{\text{TP}}{\text{TP} + \text{FN}} + \frac{\text{TN}}{\text{TN} + \text{FP}} \right) \quad (5)$$

(worst value = 0; best value = 1).

Balanced accuracy cannot, however, detect poor precision in imbalanced cases. Consider a confusion matrix where TP = 10, FN = 0, TN = 60, FP = 30. In this case, the precision is low (0.25), while the balanced accuracy is high (0.83).

The MCC takes into account all the four classes of the confusion matrix (Table 2) [19]:

$$\text{Matthews correlation coefficient (MCC)} =$$
$$\frac{\text{TP} \cdot \text{TN} - \text{FP} \cdot \text{FN}}{\sqrt{(\text{TP} + \text{FP}) \cdot (\text{TP} + \text{FN}) \cdot (\text{TN} + \text{FP}) \cdot (\text{TN} + \text{FN})}} \quad (6)$$

(worst value = −1; best value = +1).

The MCC provides a balanced measure which can be used even if the classes have different sizes [20]. A coefficient of +1.0 indicates a perfect prediction; 0.0 means the prediction is no better than random prediction; −1.0 indicates the worst prediction possible. It combines both the accuracy and the coverage of the prediction in a balanced way [21, 22]. It produces a high score only if the classifier obtained optimal results in all four of the confusion matrix cells [14, 22, 23, 24]. Researchers in machine learning and biostatistics have widely used MCC as a performance metric in assessing binary classifiers for both balanced and imbalanced datasets [25, 26, 27, 21]. While balanced accuracy fails to detect low precision, MCC successfully reflects a low positive relationship between reality and prediction.

Another metric, the $F_1$ *score*, is the harmonic mean of precision and recall:

$$F_1 = 2 \cdot \frac{\text{precision} \cdot \text{recall}}{\text{precision} + \text{recall}} = \frac{2 \cdot \text{TP}}{2 \cdot \text{TP} + \text{FP} + \text{FN}} \quad (7)$$

(worst value = 0; best value = 1).

The $F_1$ score takes both precision and recall into account at equal weights, returning values between 0.0 and 1.0. It generates a high score only if the number of true positives obtained is high compared to the other confusion matrix cells.

## 1.3 Binary classification with variable thresholds

Many classification methods generate a prediction score $f(x_i)$ between 0.0 and 1.0 inclusive for each input element $x_i$. For example, logistic regression models predict probabilities for each item in a dataset. We must therefore adopt a threshold $\tau$ to generate each confusion matrix. We classify an element as "positive" if $f(x_i) \geq \tau$, and "negative" otherwise. Varying thresholds will yield different confusion matrices and different values for each summary metric.

The ROC curve and PR curve evaluate classifiers over variable thresholds. They both plot how two single-aspect metrics relate to all possible thresholds. The ROC curve plots parametrically recall (Eqn 1) versus FPR (Eqn 2), with threshold as the varying parameter. The PR curve similarly plots parametrically precision (Eqn 3) versus recall (Eqn 1).

The *multi-threshold integration* metrics area under the ROC curve (AUROC) and area under the PR curve (AUPR) integrate classifier performance across variable thresholds (Table 2). AUROC and AUPR can range between 0 and 1. In general, larger AUROC or AUPR indicate better classifiers, but one must use some caution in interpretation. Like the ROC curve, AUROC can also mislead in the case of imbalanced ground truth [4, 5]. In PR curves, comparison between classifiers based upon area under the curve can favor models with lower $F_1$ scores [11].

Unlike the ROC and PR curves, the MCC-$F_1$ curve incorporates two total-summary metrics, MCC and the $F_1$ score. The MCC-$F_1$ curve plots parametrically the unit-normalized MCC (Eqn 6) on the y-axis and the $F_1$ score (Eqn 7) on the x-axis, with the threshold as a varying parameter.

## 2 The MCC-$F_1$ curve

To solve problems with ROC and PR analysis, we designed a more informative classification evaluation method, the *MCC-$F_1$ curve*. Our method combines the MCC and $F_1$ score, which both summarize the whole confusion matrix. ROC and PR curves, instead, combine metrics that capture only single aspect of the confusion matrix. Our method especially proves more informative when using imbalanced test datasets.

While many researchers use MCC and $F_1$ score to evaluate classifiers, most report the MCC only at particular thresholds. The MCC-$F_1$ curve, however, visualizes the MCC and $F_1$ score across the full range of possible thresholds. Therefore, it enables comparing classifiers more comprehensively than the single-threshold total summaries. It relies on MCC's capability of MCC to fairly evaluate confusion matrices with imbalanced class prevalence. This makes the MCC-$F_1$ curve more reliable than ROC and PR curves in the case of imbalanced ground truths.

### 2.1 Features and properties

The MCC-$F_1$ curve results from plotting unit-normalized MCC against the $F_1$ score with the varying parameter of threshold $\tau$. Most confusion matrix metrics range in the $[0, 1]$ interval (Table 2). MCC, instead, ranges in the $[-1, +1]$ interval. To give both axes in the MCC-$F_1$ curve the range $[0, 1]$, we rescale the MCC to the range of the $F_1$ score:

$$\text{unit-normalized MCC} = \frac{\text{MCC} + 1}{2} \tag{8}$$

(worst value = 0; best value = 1).

We highlight the key properties of the MCC-$F_1$ curve in the following paragraphs.

**Start and end points.** Usually, the MCC-$F_1$ curve starts near coordinate ($F_1$, unit-normalized MCC) = $(0, 0.5)$ with $\tau$ close to 1 and ends close to a coordinate $(k, 0.5)$ as $\tau$ decreases (Figure 1). The value of $k$ is the $F_1$ when all elements are predicted as positive. This quantity is a function of the ground truth class prevalence, as described below. In detail:

1. When threshold $\tau = 1$, all the items in the dataset are predicted to be negative (TP = 0, FP = 0). Therefore, recall, the denominator of precision, and the denominator of MCC are all 0 (recall = 0, precision = 0/0, MCC = 0/0). To avoid division by zero, we exclude this case from plotting. When $\tau$ is nearly 1, MCC and recall are close to 0. Therefore unit-normalized MCC is close to 0.5, and $F_1$ score is close to 0.

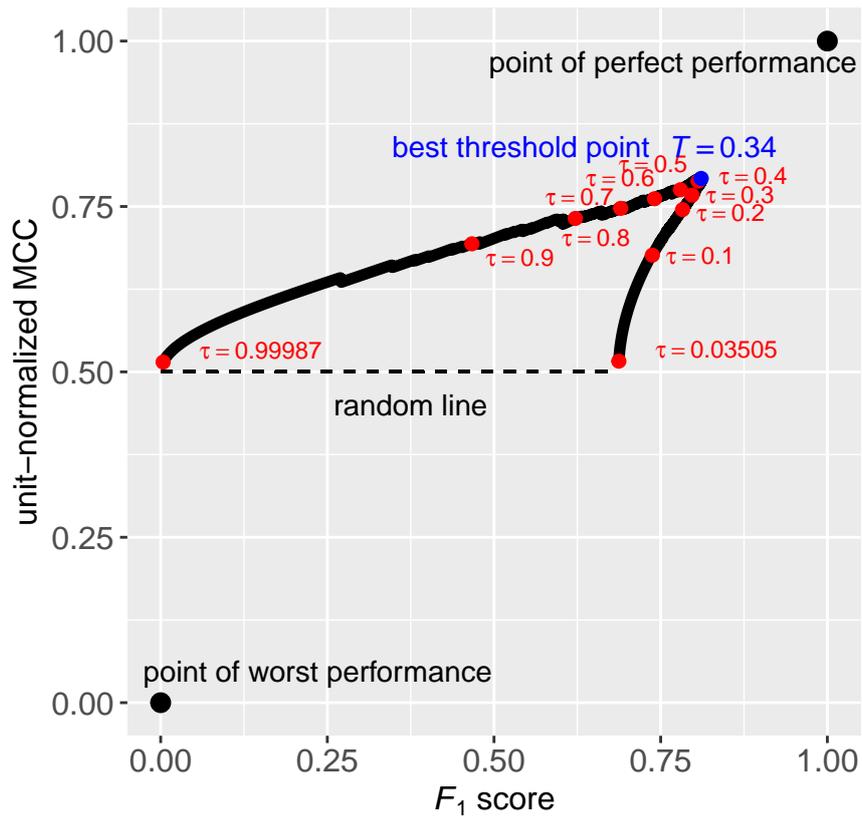

**Figure 1: Example of an MCC-F$_1$ curve.** Unit-normalized Matthews correlation coefficient (MCC) against $F_1$ score. Random line: unit-normalized MCC = 0.5, which a random classifier can achieve. Point of perfect performance: (1,1), achieved by a perfect classifier that classifies every case correctly. Point of worst performance: (0,0), achieved by the worst possible classifier, which classifies every case incorrectly. Best threshold point: point on the curve closest to (1,1).

2. Suppose the number of actual positives is P and the number of actual negatives is N. When the threshold $\tau = 0$, all the items in the dataset are predicted to be positive (TN = 0, FN = 0, TP = P, FP = N). It follows that precision = TP/(TP + FP) = P/(P + N) and recall = TP/(TP + FN) = TP/(TP + 0) = 1. We refer to the value of $F_1$ when $\tau = 0$ as $k$:

$$k = F_1 = 2 \cdot \frac{\text{precision} \cdot \text{recall}}{\text{precision} + \text{recall}} = 2 \cdot \frac{\frac{P}{P+N} \cdot 1}{\frac{P}{P+N} + 1} = 2 \cdot \frac{\frac{P}{P+N}}{\frac{P+(P+N)}{P+N}} = 2 \cdot \frac{P}{2P + N}. \tag{9}$$

It is a function only of the ground truth, irrespective of the classifier used.

When $\tau = 0$, the denominator of MCC is 0 (MCC = 0/0), so we exclude this case from plotting. When $\tau$ slightly exceeds 0, MCC is close to 0, recall is nearly 1, and precision is close to P/(P + N). Hence, the $F_1$ score is close to $k$.

**Random line, point of perfect performance, point of worst performance, and best threshold point.** A horizontal line at unit-normalized MCC = 0.5 is the *random line*, which a random classifier can achieve. Any better classifier plots above the random line. The *point of perfect performance* is (1,1), the result of a perfect classifier that predicts all data elements correctly. The point (0,0) is the *point of worst performance*, the result of a worst-case classifier that classifies all elements incorrectly (Figure 1). The *best threshold point* provides the best prediction score threshold $T$ for the confusion matrix. This allows us to select the optimal value to discriminate between false positives and true negatives, and between false negatives and true positives.

## 2.2 The MCC-$F_1$ metric

To compare classifier performance across varying thresholds, we want a metric that integrates over the MCC-$F_1$ curve. This metric should reflect that MCC-$F_1$ curves closer to the point of perfect performance indicate better classifiers. Therefore, we developed a metric that summarizes the distance between all the points on the MCC-$F_1$ curve and the point of perfect performance (1,1).

Average distance between points on the MCC-$F_1$ curve and the point of perfect performance might seem like an attractive choice for a summary metric. Unfortunately, if we use all prediction scores $f(x_i)$ generated for an input dataset, the metric displays undesirable behavior with some classifiers. Specifically, this affects classifiers that can achieve very high MCC and $F_1$ score but where the input dataset yields many prediction scores with low MCC. These cases produce a large average distance, despite the quality of the classifier.

We sought a metric that would summarize the entire MCC-$F_1$ curve while avoiding penalizing those that have many prediction scores with low MCC. Instead of averaging over all prediction scores generated for an input dataset, we consider the distance to the point of perfect performance over different ranges of MCC. We also divided the MCC-$F_1$ curve into two *sides*, the *left side* (L) and the *right side* (R). The left side contains the points with prediction scores equal to or exceeding the best prediction score threshold, $f(x_i) \geq T$. The right side contains the rest of the MCC-$F_1$ curve. Considering the left side and the right side separately avoids weighting a side with a greater number of points more. This ensures that the metric provides a balanced evaluation of the full range of prediction values.

To calculate the *MCC-$F_1$ metric*, we implemented the following procedure. First, for each of the $N$ points $i \in [0, N-1]$ on the MCC-$F_1$ curve corresponding to a prediction score $f(x_i)$, we identified the unit-normalized MCC $X_i$ and the $F_1$ score $Y_i$. Second, we divided the range of normalized MCC in the curve $[\min_i X_i, \max_i X_i]$ into $W = 100$ sub-ranges, each of width $w = (\max_i X_i - \min_i X_i)/W$. These sub-ranges cover the normalized MCC intervals $[\min_i X_i, \min_i X_i + w), [\min_i X_i + w, \min_i X_i + 2w), \ldots, [\min_i X_i + (W-1)w, \max_i X_i]$. Generally, larger values of $W$ will cause the MCC-$F_1$ metric to capture the performance of a classifier more accurately. Third, for each side, we calculated the mean Euclidean distance between points with MCC in each sub-range to the point of perfect performance. Since MCC varies non-monotonically with prediction score, we had to consider the full range of MCC on both sides. Fourth, we averaged these mean distances.

To calculate mean Euclidean distances, we began by calculating the Euclidean distance $D_i$ between each single point $i$ and the point of perfect performance (1,1) (Eqn 10):

$$D_i = \sqrt{(X_i - 1)^2 + (Y_i - 1)^2}. \qquad (10)$$

The distance between the point of worst performance (0,0) and the point of perfect performance (1,1) is $\sqrt{2}$. That is the maximum possible distance between a point on the MCC-$F_1$ curve and the point of perfect performance.

We examined each side $s \in \{L, R\}$, identifying those points whose normalized MCC resides in sub-range $j \in [0, W-1]$. For the set of such points $\mathcal{Z}_j^s$, we identified the number of points $n_j^s$ (Eqn 11):

$$n_j^s = |\mathcal{Z}_j^s|. \qquad (11)$$

When the set has a nonzero number of points, we also defined the mean distance $\bar{D}_j^s$ (Eqn 12):

$$\bar{D}_j^s = \frac{\sum_{i \in \mathcal{Z}_j^s} D_i}{n_j^s}. \qquad (12)$$

To get the grand average distance $D^*$, we began by identifying all (side, sub-range) pairs $\mathcal{P} = (s, j)$ where $\mathcal{Z}_j^s$ with a nonzero number of points (Eqn 13):

$$\mathcal{P} = \{(s, j) \mid s \in \{L, R\}, j \in [0, W-1], n_j^s > 0\}. \qquad (13)$$

We then averaged the mean distances $\bar{D}_j^s$ over these pairs to get the grand average distance (Eqn 14):

$$D^* = \frac{\sum_{(s,j) \in \mathcal{P}} \bar{D}_j^s}{|\mathcal{P}|}. \qquad (14)$$

To compare the grand average distance $D^*$ to $\sqrt{2}$, the distance between the point of worst performance (0,0) and the point of perfect performance (1,1), we took their ratio. This ratio ranges between 0 and 1. To get the MCC-$F_1$ score, we subtracted this ratio from 1 (Eqn 15):

$$\text{MCC-}F_1 \text{ metric} = 1 - \frac{D^*}{\sqrt{2}} \qquad (15)$$

(worst value = 0; best value = 1).

Better classifiers have MCC-$F_1$ curves closer to the point of perfect performance (1,1), and have a larger MCC-$F_1$ metric.

### 2.3 Confusion matrix threshold optimization

MCC-$F_1$ analyses can also find the best binary classification threshold. We consider the best threshold $T$ the point on the MCC-$F_1$ curve closest to the point of perfect performance (1,1). This threshold maximizes the overall advantage in MCC and $F_1$ score. The ROC curve and the PR curve provide no information about the best score threshold to use, as they only present the trade-offs of single-aspect metrics. Instead, the MCC-$F_1$ curve allows determining the best threshold by using two total-summary metrics.

## 3 Simulations

To compare the use of the MCC-$F_1$ curve to that of the ROC curve and the PR curve, we simulated binary classification on three datasets. The simulated datasets differ only in their ground truth class prevalence (Table 3). Dataset x skews towards the negative class, dataset y skews towards the positive class, and dataset z has perfect balance between the two classes. We represented each of the three datasets and the performance of two different simulated classifiers A and B on them as a table with three columns (Table 4). In each row $i$, the first column $y_i$ represents the ground truth (0 or 1). The second column $f_A(x_i)$ represents the prediction scores of classifier A, and the third $f_B(x_i)$ the prediction scores of classifier B.

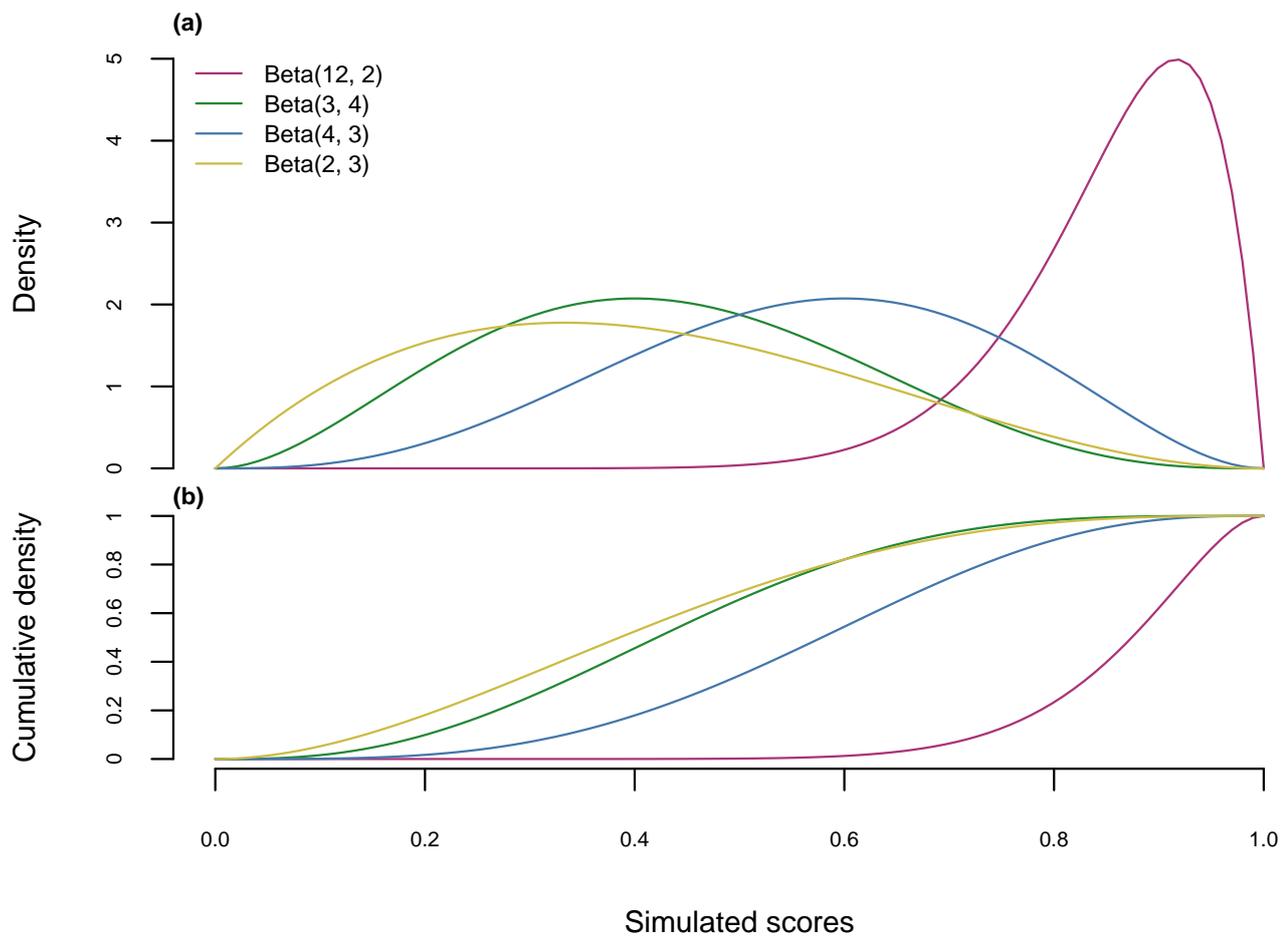

**Figure 2: Prediction scores sampled from four beta distributions.** We simulated prediction scores of six classifications by sampling from four beta distributions. By changing beta distribution shape parameters, we simulated different classifiers. (**a**) Density function of the distributions. (**b**) Cumulative density function of the distributions.

| dataset | number of actual positives | number of actual negatives |
|---|---:|---:|
| x | 1,000 | 10,000 |
| y | 10,000 | 1,000 |
| z | 10,000 | 10,000 |

Table 3: Summary statistics of the ground truths of three simulated datasets.

| ground truth | prediction scores of classifier A | prediction scores of classifier B |
|:---:|:---|:---|
| 0 | Beta(2, 3) | Beta(2, 3) |
| 1 | $\begin{cases} \text{Beta}(12, 2) & \text{for first 30\%} \\ \text{Beta}(3, 4) & \text{for last 70\%} \end{cases}$ | Beta(4, 3) |

Table 4: Score distributions used for two simulated classifiers.

To simulate the prediction scores, we sampled from the beta distribution Beta($\alpha, \beta$), a family of probability distributions with two shape parameters $\alpha$ and $\beta$ [28] (Figure 2). We used R [29] 3.3.3 for these simulations and all other analyses. The beta distribution is defined on [0,1], just like the range of prediction scores. By changing the beta distribution shape parameters, we simulated different classifiers. Higher $\alpha$ and lower $\beta$ generates a beta distribution with more density closer to 1.

In total, the three datasets and two classifiers represent six individual binary classifications: $A_x$, $B_x$, $A_y$, $B_y$, $A_z$, and $B_z$. Dataset x's ground truth column contains 1,000 actual positives and 10,000 actual negatives. The dataset has 11,000 corresponding prediction scores in classification $A_x$ and 11,000 scores in classification $B_x$. Dataset y contains 10,000 actual positives and 1,000 actual negatives, leading to 11,000 prediction scores for each of $A_y$ and $B_y$. Dataset z contains 10,000 actual positives and 10,000 actual negatives, leading to 20,000 prediction scores for each of $A_z$ and $B_z$.

To simulate prediction scores corresponding to particular ground truth values for classifier A, we sampled from the same beta distributions across all datasets (Table 4). For each dataset, we sampled from Beta(12, 2) to simulate prediction scores of classifier A for the first 30% of actual positives. For the remaining actual positives, we sampled from Beta(3, 4) instead (Table 4). This bimodal piecewise distribution simulates a classifier with large recall when the threshold nears 1 and small recall when the threshold nears 0. For actual negatives, we sampled the prediction scores of classifier A from Beta(2, 3).

For classifier B, we sampled from other beta distributions across all datasets (Table 4). For all actual positives, we sampled from Beta(4, 3). For all actual negatives, we sampled from Beta(2, 3) just like classifier A.

Each classifier can produce different results when applied to datasets with different class prevalences (Figure 3). For example, when threshold $\tau = 0.5$, classifier A exhibits the largest precision with dataset y (0.95), followed by dataset z (0.63), and then dataset x (0.15).

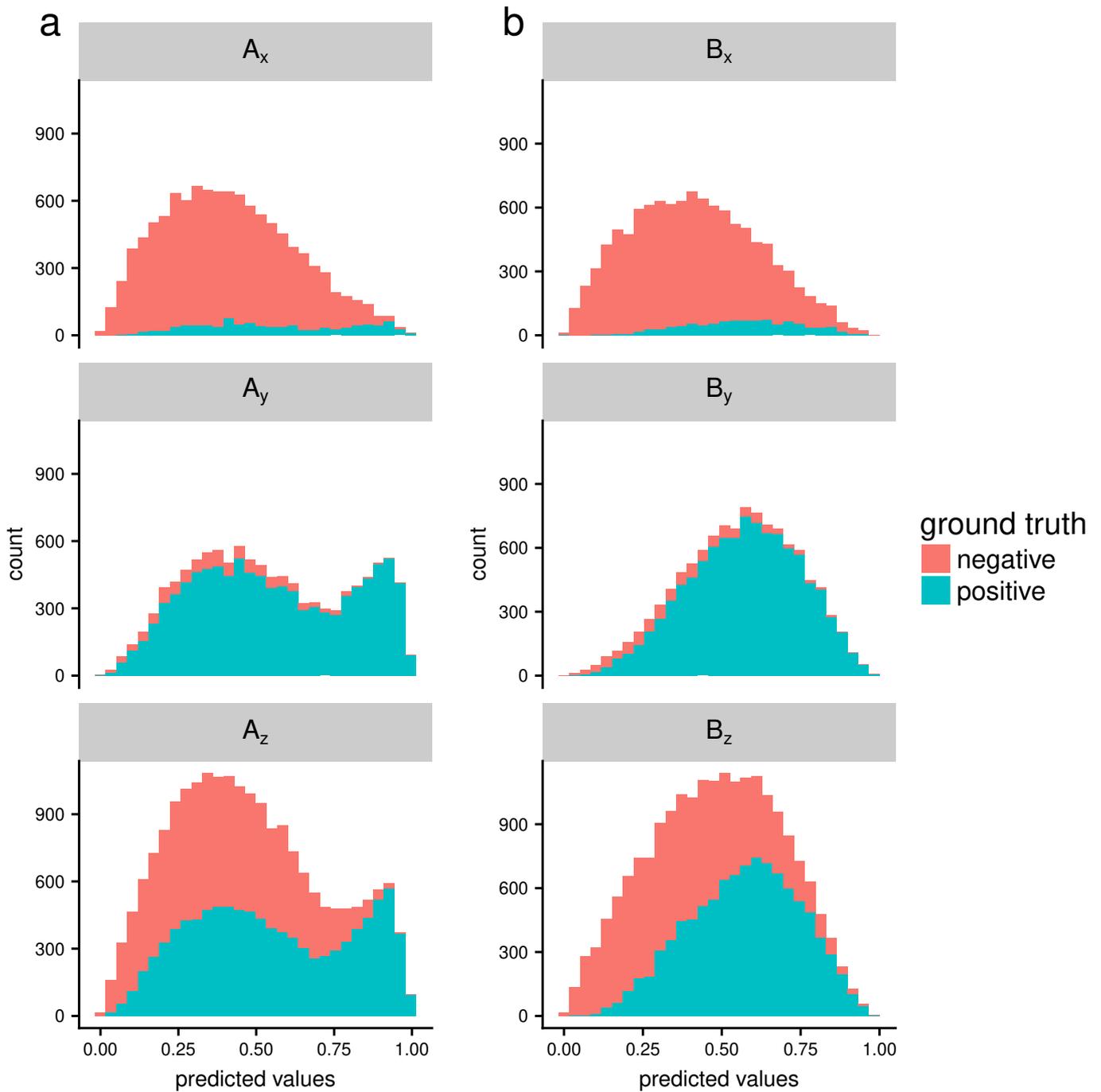

**Figure 3: Histograms of predicted values of classifications $A_x$, $B_x$, $A_y$, $B_y$, $A_z$, and $B_z$.** Each row has histograms of prediction scores from one dataset: x (1,000 actual positives, 10,000 actual negatives), y (10,000 actual positives, 1,000 actual negatives), and z (10,000 actual positives, 10,000 actual negatives). Histograms of predicted values of (**a**) classifications $A_x$, $A_y$, and $A_z$, and (**b**) classifications $B_x$, $B_y$, and $B_z$

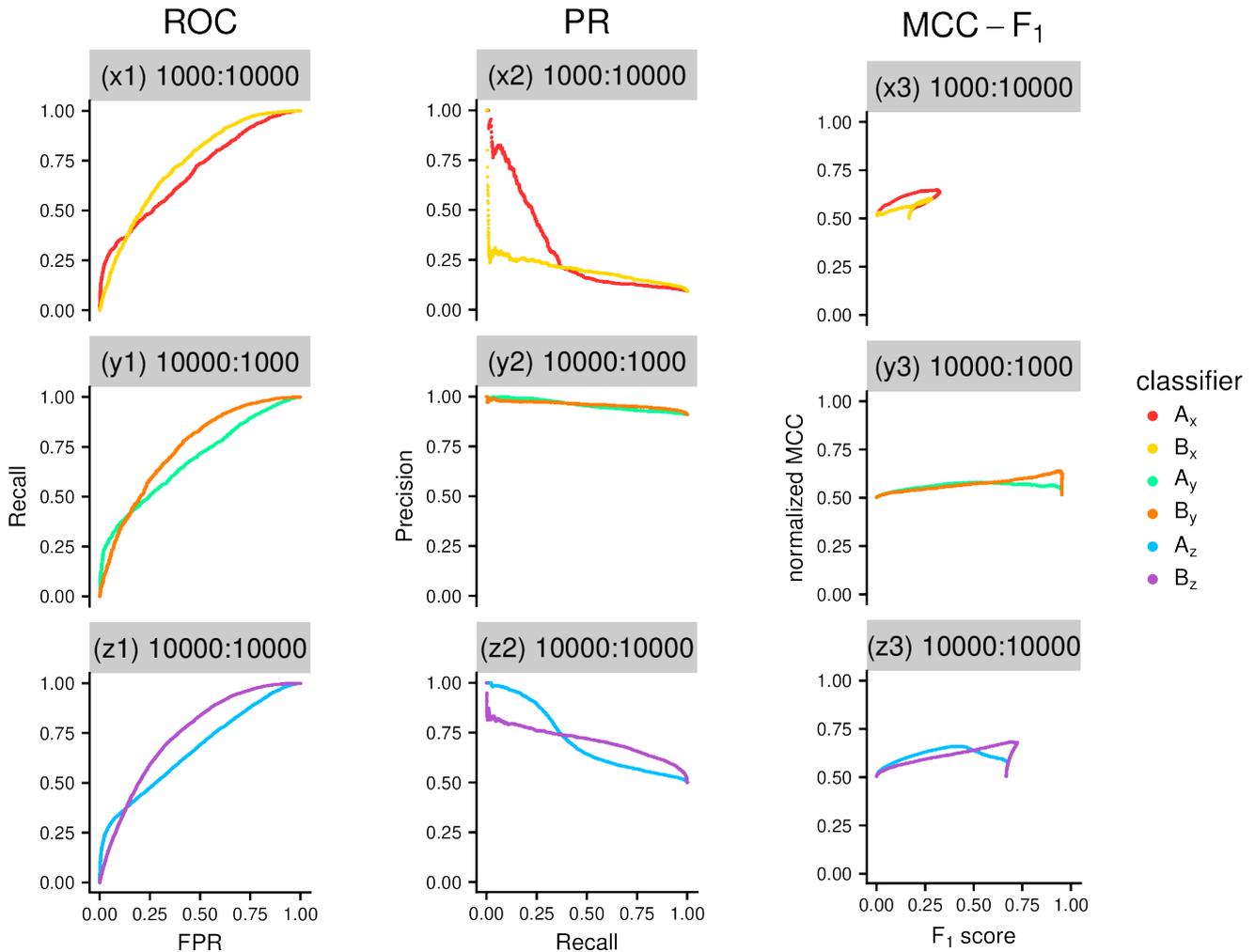

**Figure 4: ROC, PR, and MCC-$F_1$ curves of classifications $A_x$, $B_x$, $A_y$, $B_y$, $A_z$, and $B_z$** Each column presents a benchmark curve for a set of classifiers (left: ROC, center: PR, right: MCC-$F_1$). **(x1–x3)** dataset x, negatively imbalanced with 1,000 actual positives and 10,000 actual negatives. **(y1–y3)** dataset y, positively imbalanced with 10,000 actual positives and 1,000 actual negatives. **(z1–z3)** dataset z, balanced with 10,000 actual positives and 10,000 actual negatives.

# 4 Results

## 4.1 MCC-$F_1$ analysis is more informative than ROC analysis on a negatively imbalanced dataset

Imbalanced datasets skewed towards the negative class occur commonly in many research fields. Such datasets have more actual negative data instances than actual positives. Some examples include genomic data used to predict transcription factor presence (genomic regions with transcription factors present: 16,559, regions without: 2,650,396) [30], personal finance data used to predict whether individuals earn over $50,000 a year (people who earn > $50,000: 12,103, number of people who earn < $50,000: 36,739) [31, 32], and mammography data used to predict tumor severity (malignant abnormalities: 435, benign abnormalities: 65,365) [6, 32].

Dataset x has imbalanced class prevalance (Figure 4x1–x3) and demonstrates common problems of ROC analysis [6, 7, 33]. ROC curves can present an overly optimistic view of a classifier's performance, in the case of large imbalance [7].

ROC analysis shows that classification $B_x$ has better performance than classification $A_x$ except when FPR < 0.12 (Figure 4x1). Indeed, the AUROC of classification $B_x$, 0.73, exceeds the AUROC of classification $A_x$, 0.69 (Table 5). With PR analysis, however, classification $A_x$ outper-forms classification $B_x$ with substantial advantage when recall is less than 0.4 (Figure 4x2). While classification $B_x$ outperforms classification $A_x$ when recall exceeds 0.4, the difference in precision between the two classifiers never exceeds 0.05 at fixed recall. Classification $B_x$, nearly dominant in the ROC analysis, loses its advantages in the PR analysis.

| dataset | classification | AUROC | AUPR | MCC-$F_1$ metric |
|---|---|---|---|---|
| x | $A_x$ | 0.69 | **0.30** | **0.35** |
| x | $B_x$ | **0.73** | 0.20 | 0.34 |
| y | $A_y$ | 0.69 | 0.96 | 0.46 |
| y | $B_y$ | **0.73** | 0.96 | **0.59** |
| z | $A_z$ | 0.69 | 0.71 | 0.46 |
| z | $B_z$ | **0.73** | 0.71 | **0.53** |

Table 5: **Performance of classifications $A_x$, $B_x$, $A_y$, $B_y$, $A_z$, and $B_z$ evaluated with 3 multi-threshold integration metrics.** Bold: the best classification for each (dataset, metric) pair. For ties, neither is marked bold.

In a negatively imbalanced dataset, large changes in the number of false positives can lead to only small changes in the FPR. ROC curves therefore present an overly optimistic view of classification $A_x$. When the two classifiers have the same recall < 0.4, they both produce a small number of true positives and classification $A_x$ outperforms classification $B_x$. Because of the large number of actual negatives, the small difference in FPR (Figure 4x1) counterintuitively indicates a large difference in the number of false positives.

Since precision = TP/(TP+FP), if we hold fixed a small number of true positives, a large change in the number of false positives will yield a large change in precision. This explains why in the PR analysis, classification $A_x$ outperforms classification $B_x$ significantly when recall < 0.4 (Figure 4x2). When recall exceeds 0.4, the difference between the two classifiers' precision is massive for two reasons. First, the number of true positives is relatively large with large fixed recall. Second, the number of false positives are also relatively large. Although the number of false positives changes greatly, this has small impact on the difference in precision, because the number of true positives and false positives are also large.

Classification $B_x$, seemingly satisfactory in ROC analysis (AUROC: 0.73), performs badly in PR analysis with precision generally below 0.5. In this scenario, using ROC analysis alone misleads. PR analysis represents the poor precision of classification $B_x$, and here proves more informative than ROC analysis.

In MCC-$F_1$ analysis, classification $B_x$ performs worse than classification $A_x$. Classification $B_x$'s MCC-$F_1$ curve is further away from the point of perfect performance than classification $A_x$ (Figure 4x3). The MCC-$F_1$ metric (Eqn 15) of classification $A_x$ is 0.35. This score exceeds that of classification $B_x$ (section 4). The MCC-$F_1$ curve clearly shows that classification $B_x$ is not as good as the ROC curve misleadingly suggests.

As an additional benefit, MCC-$F_1$ analysis determines the best threshold to use. Using the MCC-$F_1$ metric, the best threshold for classification $A_x$ is 0.79 and the best threshold for classification $B_x$ is 0.60.

## 4.2 MCC-$F_1$ analysis is more informative than PR analysis on a positively imbalanced dataset

Imbalanced datasets skewed towards the positive class often appear in genomics and biomedicine. Often one cannot identify a complete set of actual negatives in genomic classification, or one has a health dataset where sick patients (labeled positive) represent the majority. Lian and colleagues [34], for example, examined subjects with esophageal cancer symptoms with a dataset of 13 disease-free individuals (labeled negative) and 23 sick patients (labeled positive).

Dataset y (Figure 4y1,y3) illustrates a typical scenario where classifications have nearby PR

curves due to positive class imbalance. PR analysis proves ineffective when the number of actual positives exceeds the number of actual negatives. Most of the recall domain corresponds to a confusion matrix with a large number of true positives. This arises from the large number of actual positives in the dataset. The change in precision, therefore, would remain small even if the number of false positives changed substantially. Therefore, classification $A_y$ and classification $B_y$ have similar PR curves (Figure 4y2). Both classifications have an AUPR of 0.96 (Table 5). Here, ROC analysis more clearly indicates that classification $B_y$ outperforms classification $A_y$ (Figure 4y1). Classification $B_y$ has obvious advantage in most of the FPR domain.

Classification $B_y$ and classification $A_y$ have nearby MCC-$F_1$ curves when $F_1 < 0.62$ (Figure 4y3). When $F_1 > 0.62$, MCC-$F_1$ analysis shows that classification $B_y$ performs better than classification $A_y$. The MCC-$F_1$ metric of classification $B_y$ (0.59) exceeds that of classification $A_y$ (0.46). Here, we cannot differentiate the two classifiers by PR analysis. ROC and MCC-$F_1$ analyses, however, both show that classification $B_y$ performs better overall. In this case, both ROC and MCC-$F_1$ analyses prove more informative than PR analysis. Additionally, the MCC-$F_1$ metric tells us the best threshold for both classification $A_y$ (0.22) and for classification $B_y$ (0.26).

## 4.3 MCC-$F_1$ analysis provides a clear decision rule for picking a classifier and threshold on a balanced dataset

The perfectly balanced dataset z provides an ambiguous scenario where both the ROC and the PR curves of classifiers A and B cross (Figure 4z1,z2). This makes it difficult to select the best classifier. The ROC curve suggests that classification $B_z$ performs better than classification $A_z$ (Figure 4z1). When recall is less than 0.4, however, the PR curve of classification $A_z$ outperforms the PR curve of classification $B_z$ with great advantage (Figure 4z2).

For dataset z, the two MCC-$F_1$ curves also cross, therefore producing an ambiguous outcome (Figure 4z3). The MCC-$F_1$ metric tells us that classification $B_z$ generally performs better (0.53) than classification $A_z$ (0.46). Nonetheless, MCC-$F_1$ analysis does not indicate which classifier performs best in all cases. ROC and PR analysis, however, cannot do this either. At least, the MCC-$F_1$ metric can provide a clear decision rule for picking a classifier even in this case (Table 5). And again, the MCC-$F_1$ metric provides a clear rule for picking the threshold as well.

## 5 Discussion

MCC-$F_1$ analysis compares classifiers more clearly than ROC or PR analysis. As shown above, imbalanced datasets can cause ROC or PR analyses to lead to misleading conclusions. The MCC-$F_1$ curve largely solves these problems, by providing clearer, more consistent way to evaluating binary classifiers. The MCC-$F_1$ metric also provides a useful means to identify the best threshold for classification. To facilitate the broader use of the MCC-$F_1$ curve, we also provide the R package, mccf1, available on Comprehensive R Archive Network (CRAN) at https://cran.r-project.org/package=mccf1.

## Competing interests

The authors declare they have no competing interests.

## Software availability

Our R mccf1 package is available on CRAN at https://cran.r-project.org/package=mccf1. Source code is available under the GNU General Public License (GPL) version 2 at https://github.com/hoffmangroup/mccf1/. The version used for this paper is at https://cran.r-project.org/src/contrib/Archive/mccf1/mccf1_1.0.tar.gz.

# Acknowledgments

The authors thank Lei Sun (University of Toronto, ORCID: 0000-0002-5640-937X) and Robert M. Flight (University of Kentucky, ORCID: 0000-0001-8141-7788) for their help and suggestions. This work was supported by the Natural Sciences and Engineering Research Council of Canada (RGPIN-2015-03948) to Michael M. Hoffman.

# Authors' details

Chang Cao (ORCID: 0000-0002-2337-9741) is at the Statistics Program, University of Toronto, and at the Princess Margaret Cancer Centre, Toronto, Ontario, Canada.

Davide Chicco (ORCID: 0000-0001-9655-7142) was at the Princess Margaret Cancer Centre and currently is at the Krembil Research Institute, Toronto, Ontario, Canada.

Michael M. Hoffman (ORCID: 0000-0002-4517-1562) is at the Princess Margaret Cancer Centre, at the Department of Medical Biophysics and the Department of Computer Science, University of Toronto, and at the Vector Institute, Toronto, Ontario, Canada.